\documentclass{article} 
\usepackage{collas2024_conference,times}
\usepackage{easyReview}
\usepackage{subfigure}
\usepackage{booktabs}
\usepackage{multirow}
\usepackage{multicol}
\usepackage{graphicx}
\usepackage{algorithm}
\usepackage{algpseudocode}
\newcommand\Tstrut{\rule{0pt}{2.2ex}}         

\usepackage{amsmath,amsfonts,bm}

\def\eqref#1{equation~\ref{#1}}

\def\1{\bm{1}}

\def\rr{{\textnormal{r}}}

\def\ry{{\textnormal{y}}}

\def\rvh{{\mathbf{h}}}

\def\rvx{{\mathbf{x}}}
\def\rvy{{\mathbf{y}}}

\def\vt{{\bm{t}}}

\def\vw{{\bm{w}}}

\def\mI{{\bm{I}}}

\def\mU{{\bm{U}}}
\def\mV{{\bm{V}}}
\def\mW{{\bm{W}}}

\DeclareMathAlphabet{\mathsfit}{\encodingdefault}{\sfdefault}{m}{sl}
\SetMathAlphabet{\mathsfit}{bold}{\encodingdefault}{\sfdefault}{bx}{n}

\newcommand{\R}{\mathbb{R}}

\usepackage{hyperref}
\hypersetup{
    colorlinks=true,
    linkcolor=red,
    filecolor=magenta,
    urlcolor=blue,
    citecolor=purple,
    pdftitle={Overleaf Example},
    pdfpagemode=FullScreen,
    }

\title{Fixed Random Classifier Rearrangement for Continual Learning}

\author{Shengyang Huang \\
School of Information and Communication \\
Guilin University of Electronic Technology \\
\texttt{hsy\_HG97@outlook.com} \\
\And 
Jianwen Mo \\
School of Information and Communication \\
Guilin University of Electronic Technology \\
\texttt{jwmo@guet.edu.cn} \\
}

\preprintcopy
\begin{document}

\maketitle

\begin{abstract}
With the explosive growth of data, continual learning capability is increasingly important for neural networks. Due to catastrophic forgetting, neural networks inevitably forget the knowledge of old tasks after learning new ones. In visual classification scenario, a common practice of alleviating the forgetting is to constrain the backbone. However, the impact of classifiers is underestimated. In this paper, we analyze the variation of model predictions in sequential binary classification tasks and find that the norm of the equivalent one-class classifiers significantly affects the forgetting level. Based on this conclusion, we propose a two-stage continual learning algorithm named Fixed Random Classifier Rearrangement (FRCR). In first stage, FRCR replaces the learnable classifiers with fixed random classifiers, constraining the norm of the equivalent one-class classifiers without affecting the performance of the network. In second stage, FRCR rearranges the entries of new classifiers to implicitly reduce the drift of old latent representations. The experimental results on multiple datasets show that FRCR significantly mitigates the model forgetting; subsequent experimental analyses further validate the effectiveness of the algorithm.
\end{abstract}

\section{Introduction}

Continual learning refers to an ability that a model learns new knowledge without forgetting old knowledge, which is essential to true artificial intelligence. Neural Network, the representative work of modern artificial intelligence, has no continual learning ability because of catastrophic forgetting\citep{RN1, RN2, RN3, RN4, RN5}. Catastrophic forgetting refers to a phenomenon that a neural network suffers a significant performance degradation on the previous task after training with a new dataset. Many excellent works have been proposed to overcome catastrophic forgetting, such as replay methods\citep{RN6, RN7, RN8}, parameter isolation methods\citep{RN9, RN10, RN11} and regularization-based methods\citep{RN12, RN13, RN14}. The essence of all these works is to constrain the backbone of neural networks, reducing the variation of the prediction on the old task—it means, in visual classification scenario, maintaining the classification accuracy of the old task after training on the new dataset. In fact, the variation of the prediction is determined jointly by the backbone and the classifier, but the impact of the classifier is underestimated.

In this paper, we try to solve a problem: if the forgetting caused by varying backbone is unavoidable, how can we mitigate its effect by regularizing the classifier? The idea of regularizing the classifier originate from a phenomenon: given multiple binary classification tasks that sharing a same backbone, some tasks always forget more knowledge than the other; after reordering these tasks, it still happens. To explore this phenomenon, we simulate the variation of the backbone by adding a noise in the latent representation of the final layer and find that the norm of the equivalent one-class classifiers significantly affects the forgetting level. Based on this conclusion, we propose a two-stage continual learning algorithm named Fixed Random Classifier Rearrangement (FRCR). In first stage, FRCR replaces the learnable classifiers with fixed random classifiers, constraining the norm of the equivalent one-class classifiers without affecting the performance of the network. In second stage, FRCR rearranges the entries of new classifiers to implicitly reduce the drift of old latent representations. The experimental results on multiple datasets show that FRCR significantly mitigates the model forgetting; subsequent experimental analyses further validate the effectiveness of the algorithm.

\section{Related works}

\textbf{Continual learning.} The goal of continual learning can be summarized as ensuring no degradation in performance on the old task after the model trains on the new task, provided that the dataset of old tasks is unavailable. Continual learning algorithms can be categorized into three classes: replay methods, parameter isolation methods, and regularization-based methods\citep{RN15}.Replay methods mix the samples or pseudo-samples from old datasets into the dataset of the new task to retain the model's performance on the old task\citep{RN16, RN17}. Parameter isolation methods train different tasks with different network parameters to avoid the interference between old tasks and the new task\citep{RN18, RN19, RN20, RN21}. The regularization-based methods add a regularization term in the loss function during training the new task to suppress the variation of the parameters that are highly correlated with the old task\citep{RN22, RN23, RN24}.

Except for three types of methods mentioned above, some new methods have emerged in recent years\citep{RN25, RN26, RN27}. Stable-SGD\citep{RN26} investigates the several factors affecting continual learning, including batch size and learning rate decay, then proposes a naive continual learning algorithm without using replay samples, parameter isolation, and regularization terms. The proposed method in this paper is similar to stable-SGD\citep{RN26}; without saving the old dataset, expanding the network capacity, or preserving the model parameters, it mitigates the forgetting just by freezing and rearranging the classifiers. To evaluate the performance of the proposed method, it is compared with stable-SGD\citep{RN26} and EWC\citep{RN22} in the experiment section.

\textbf{Fixed classifier.} Although learnable classifiers are common for neural networks on classification tasks, some work using fixed classifiers has also appeared in recent years\citep{RN28, RN29, RN30}. \citet{RN28} directly computed the optimal structure of a classifier based on the ETF framework\citep{RN31} and used this optimal structure as a fixed classifier. \citet {RN30} designed a fixed classifier based on a d-Simplex regular polytope and used it for class incremental scenarios. Unlike these works, this paper does not use any theoretical assumptions in designing the fixed classifier, and simply achieves continual learning by rearranging the entries of the classifier based on the original initialized classifier.

\section{Preliminaries}

Given multiple binary classification tasks, consider a three-layer MLP $\displaystyle f(\rvx; \,\mU, \mV, \mW)$, suppose the input vector $\displaystyle \rvx \in \R^{m}$, the output vector $\displaystyle \rvy \in \R^{2}$. The forward propagation follows:
\begin{equation}\label{eq1}
    \rvh_1 = g \left(\mU \rvx \right)
\end{equation}
\begin{equation}\label{eq2}
    \rvh_2 = g \left(\mV \rvh_1 \right)
\end{equation}
\begin{equation}\label{eq3}
    \rvy = g \left(\mW \rvh_2 \right)
\end{equation}
where $\displaystyle \mU \in \R^{n_1 \times m}, \mV \in \R^{n_2 \times n_1}, \mW \in \R^{2 \times n_2}, \rvh_1 \in \R^{n_1}, \rvh_2 \in \R^{n_2}, \rvy \in \R^{2}.$ $\displaystyle g(\cdot)$ represents the ReLU activation function. This paper focuses on the multi-head classifiers continual learning scenario, so MLP is divided into the backbone including $\displaystyle \mU, \mV$ and the classifier $\displaystyle \mW$, which means $\displaystyle \mU, \mV$ are shared by all tasks while the classifier $\displaystyle \mW$ is exclusive to only one task. To be specific, $\displaystyle \mW$ will be frozen after the current task finishes and we initialize a new classifier for the new task.

\subsection{The norm growth of equivalent one-class classifiers}

Consider many binary classification tasks, let $\displaystyle \mW = \left[\vw_1 \, \vw_2 \right]^\top$, $\displaystyle \rvy = \left[\ry_1 \, \ry_2\right]^\top$. In classification continual learning scenario, the extent of forgetfulness of the model is always measured by the variation of classification accuracy before and after the current task, which is non-intuitive and difficult to quantitatively analyze. For simpler analysis, we omit the common classification network discriminant rule based on Softmax function and choose a simple rule:
\begin{equation}\label{eq4}
    \rvx \in \left\{\begin{matrix}
                    c_1, \text{if} \, \ry_1-\ry_2 > 0\\
                    c_2, \text{if} \, \ry_1-\ry_2 < 0
                    \end{matrix}\right.
\end{equation}
where $\displaystyle c_1$ represents class 1, $\displaystyle c_2$ represents class 2. Then Eq.(\ref{eq3}) can be replaced by:
\begin{equation}\label{eq5}
\hat{\ry} = \hat{\vw} \rvh_2
\end{equation}
where $\displaystyle \hat{\vw}=\vw_1-\vw_2, \hat{\ry}=\ry_1-\ry_2.$ For simplicity, $\displaystyle \hat{\vw} \in \R^{n_2}$ and $\displaystyle \hat{\ry} \in \R$ are named as the equivalent one-class classifier and the equivalent prediction respectively. In this rule, the extent of forgetfulness can be easily described by the equivalent prediction. Assume that there are two tasks, task A and task B. The parameter update of the backbone in task B will drift the task A prediction  and the model appears to be forgotten. The forgetfulness can be approximated by the difference of the equivalent prediction:
\begin{equation}\label{eq6}
\left| \Delta\hat{\ry}^{\text{A}} \right| = \left| \hat{\ry}^{\text{A}(\vt_\text{B})} - \hat{\ry}^{\text{A}(\vt_\text{A})} \right|
\end{equation}
where $\displaystyle {\hat{\ry}^{\text{A}(\vt_\text{B})}}$ and 
$\displaystyle {\hat{\ry}^{\text{A}(\vt_\text{A})}}$ are the equivalent predictions of dataset A after task A and task B respectively, $\displaystyle \left| \Delta\hat{\ry}^{\text{A}} \right|$ is the equivalent prediction offset which reflects the stability of the network. To simulate the latent representations drift caused by the backbone, we added a noise to $\displaystyle \rvh_2^{\text{A}(\vt_\text{A})}$:
\begin{equation}\label{eq7}
\rvh_2^{\text{A}(\vt_\text{B})} = \rvh_2^{\text{A}(\vt_\text{A})} + k\epsilon
\end{equation}
Substituting Eq.(\ref{eq7}) into Eq.(\ref{eq6}):
\begin{equation}\label{eq8}
||\Delta\hat{\ry}^{\text{A}}|| = k||\epsilon \cdot \hat{\vw}||
\end{equation}
where noise $\displaystyle \epsilon \sim \mathcal{N} (\mathbf{0}, \mI), k$ is the noise amplitude.
\begin{figure}[t]
    \centering
    \begin{subfigure}[The accuracy in normal task order]
        {\includegraphics[width=0.4\textwidth]{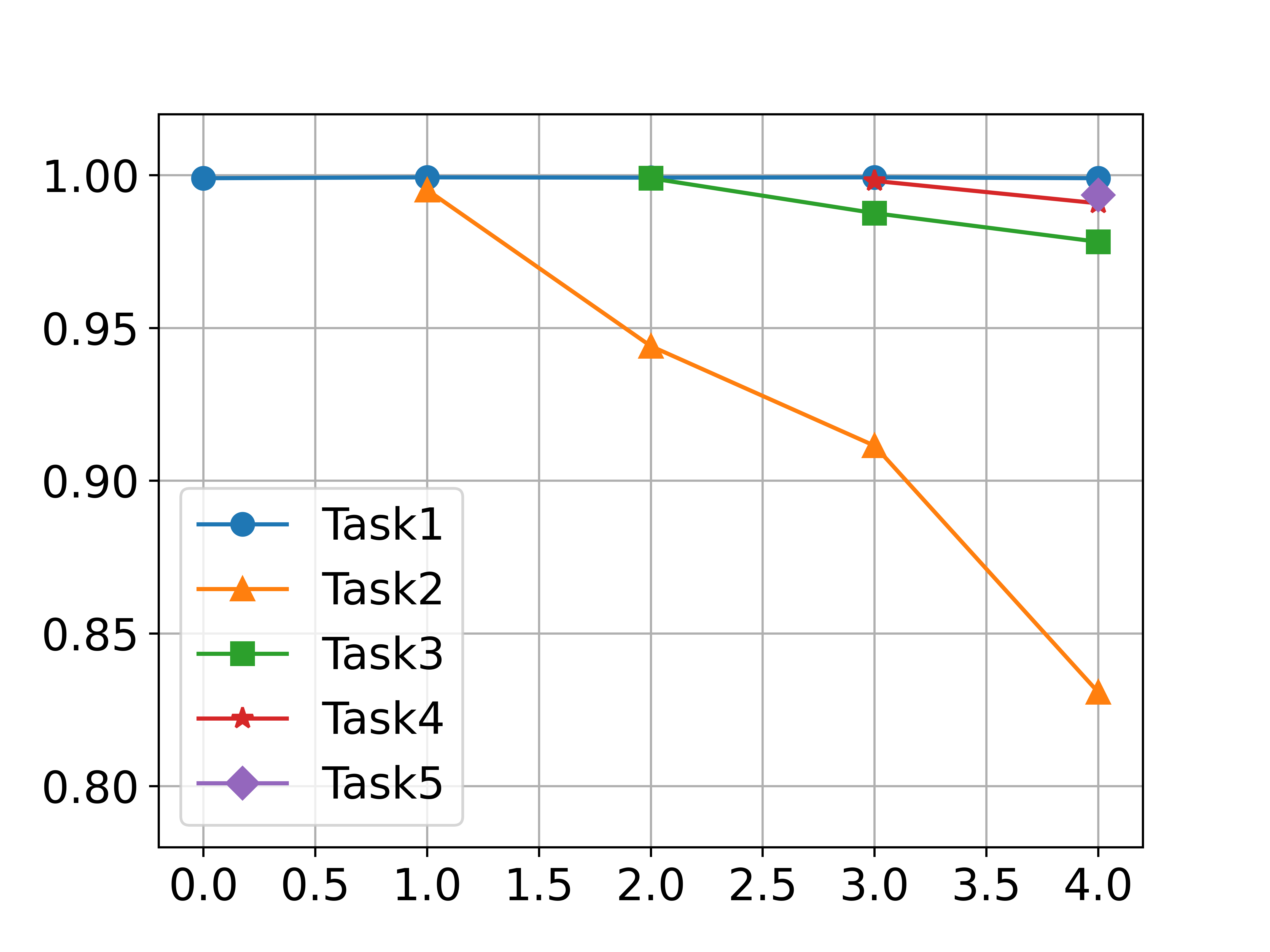}
        \label{fig1:subfig1}}
    \end{subfigure}
    \begin{subfigure}[The accuracy after reordering tasks]
        {\includegraphics[width=0.4\textwidth]{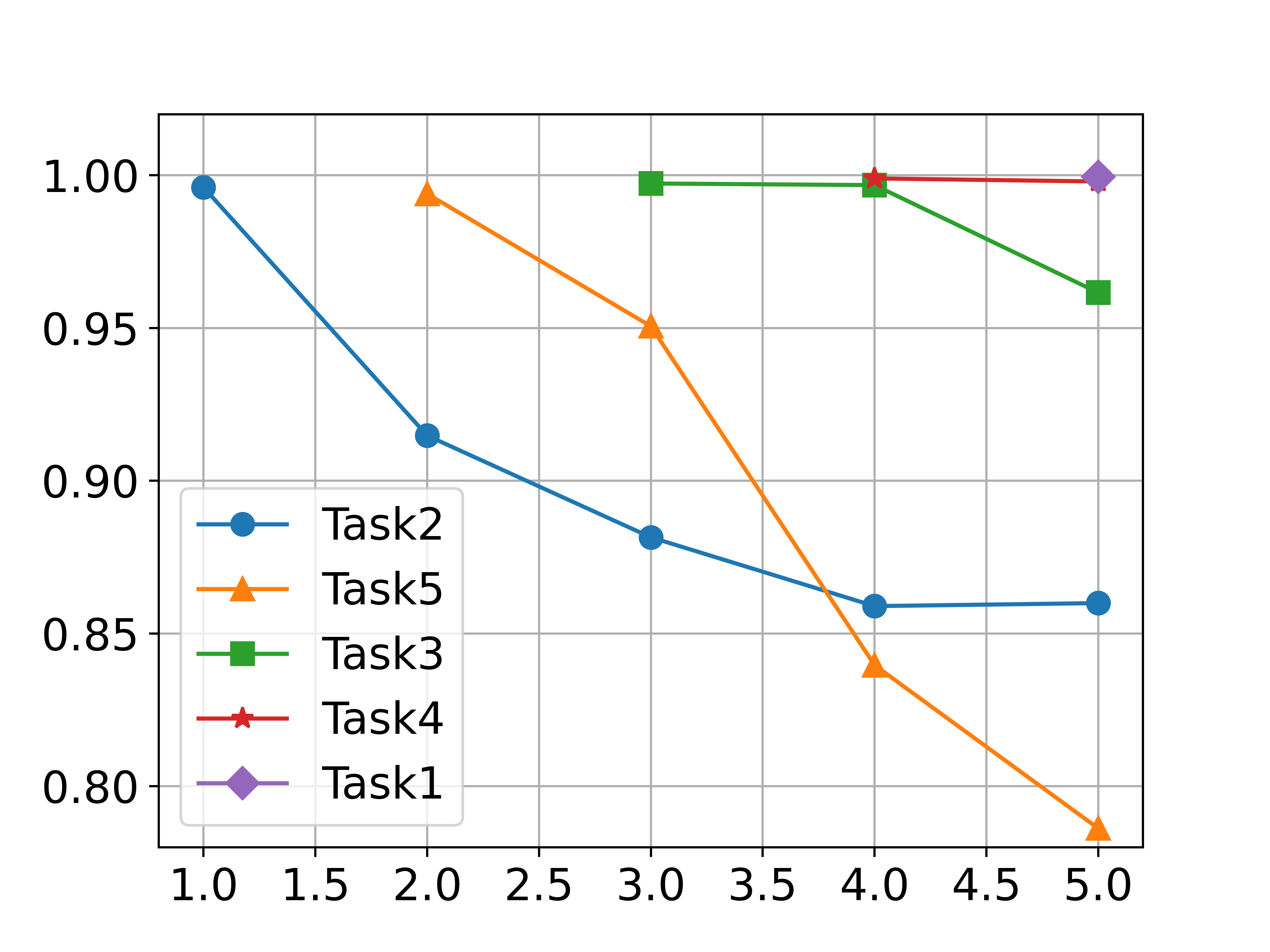}
        \label{fig1:subfig2}}
    \end{subfigure}
    \begin{subfigure}[Relationship between $\displaystyle \Delta\hat{\ry}^{\text{A}}$ and noise amplitude $\displaystyle k$ for different tasks]
        {\includegraphics[width=0.4\textwidth]{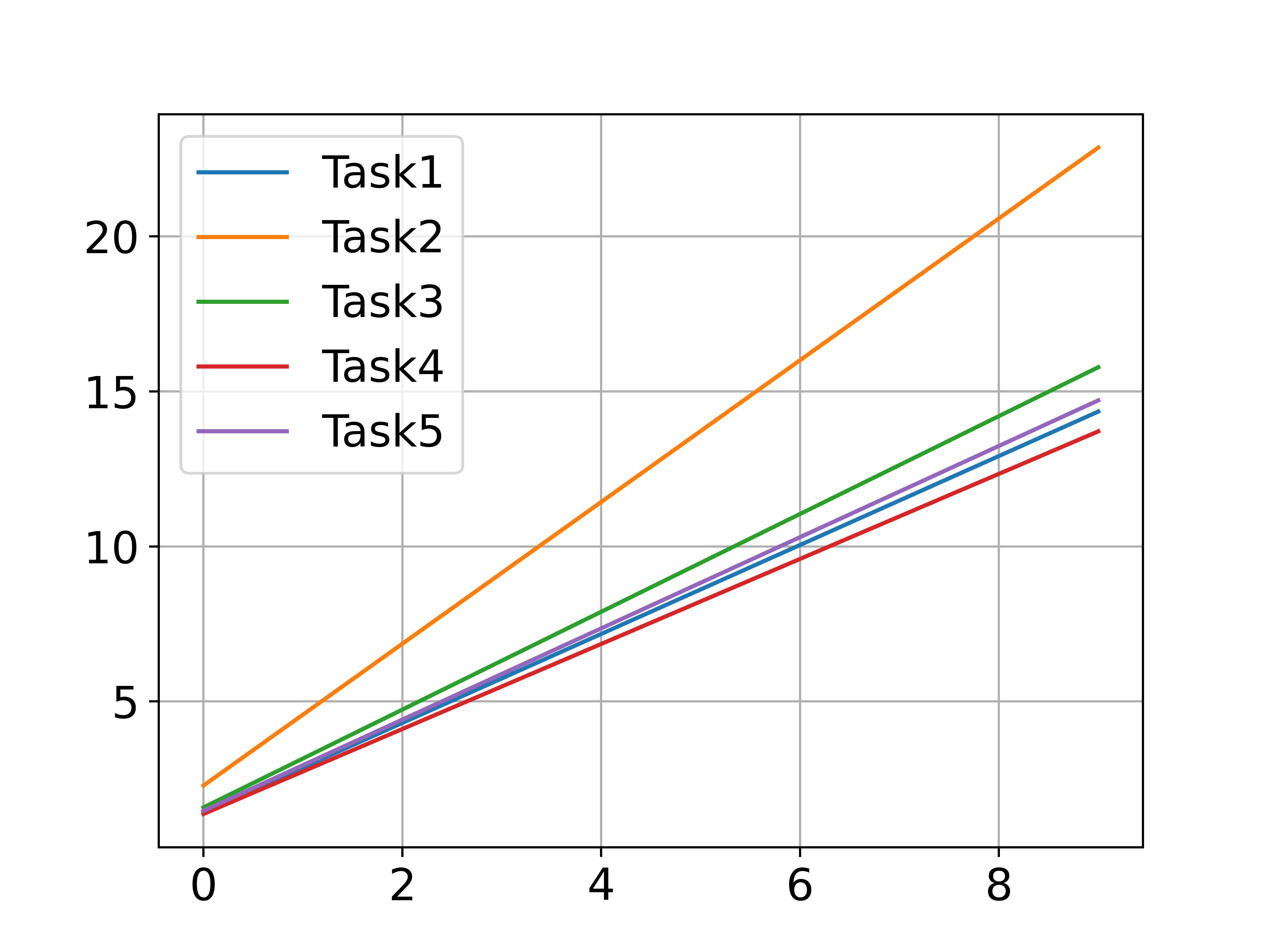}
        \label{fig1:subfig3}}
    \end{subfigure}
    \begin{subfigure}[The training trajectory of $\displaystyle \hat{\vw}^{\text{A}(\vt)}$]
        {\includegraphics[width=0.4\textwidth]{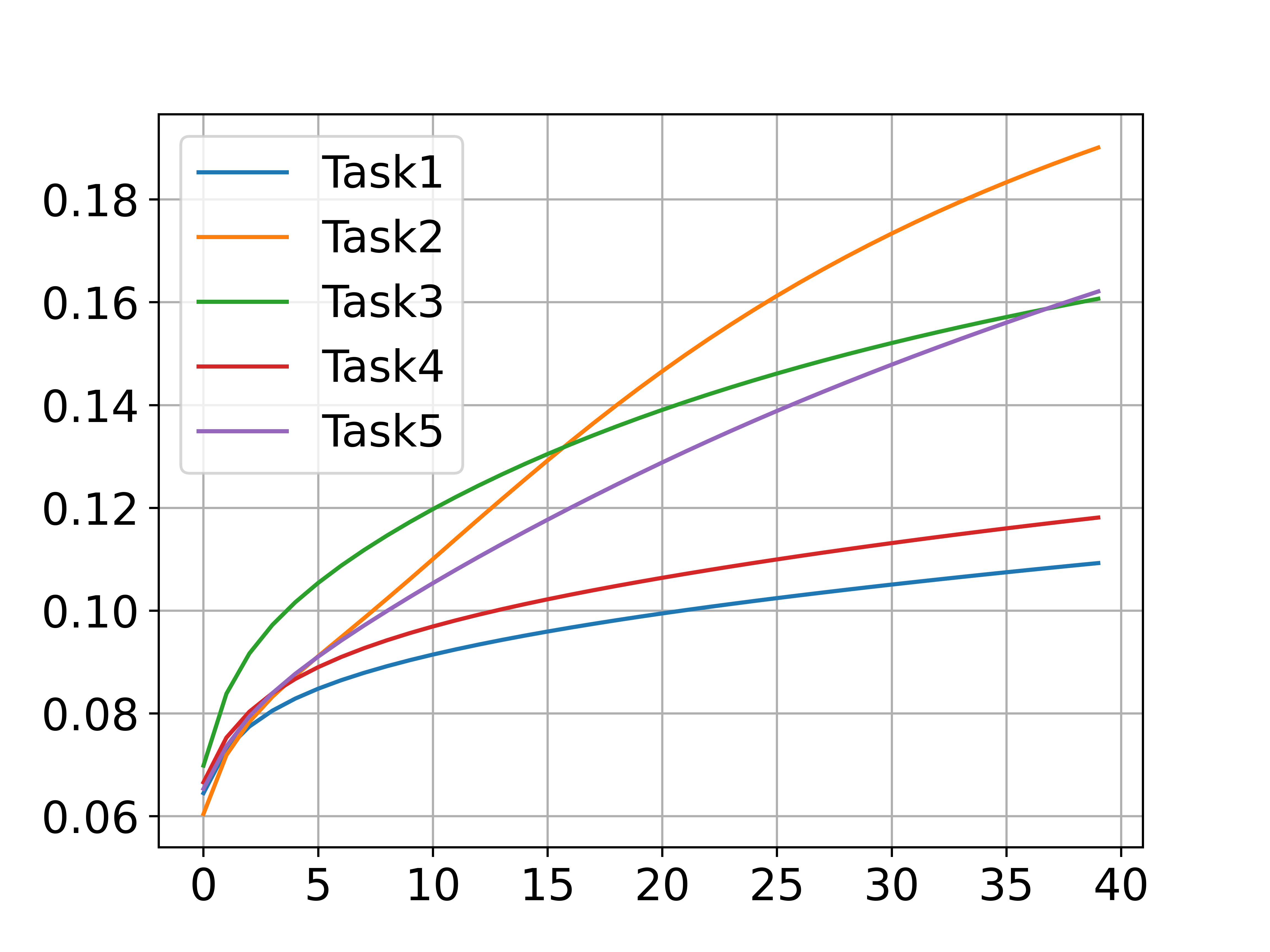}
        \label{fig1:subfig4}}
    \end{subfigure}
    \caption{The accuracy and trajectories}
    \label{fig1}
\end{figure}
Fig.\ref{fig1:subfig3} shows the relationship between $\displaystyle \Delta\hat{\ry}^{\text{A}}$ and noise amplitude $\displaystyle k$ under different tasks. The variation of the equivalent prediction of task 2 is significantly higher than that of other tasks, which is consistent with the high forgetting rate of task 2 in Fig.\ref{fig1:subfig1}, Fig.\ref{fig1:subfig2}. Fig.\ref{fig1:subfig4}) shows the norm of the equivalent one-class classifier under the number of iterations. Comparing the final results of each task, the norm of task 2 is significantly higher than that of other tasks. These results indicates that the equivalent classifier with high norm will magnify the variation of latent representation in old datasets, leading to more serious forgetfulness.

\section{Methods}
\subsection{Fixed Random Classifier and Parameter Rearrangement Algorithm}
To constrain the norm of classifier, we propose a fixed random classifiers algorithm, which freezes the classifier after random initialization. Because the norm of classifier will keep increasing in training process\citep{RN32}, it’s difficult to constraint the norm by normalizing the classifier during training dynamics. Fixing the classifier is a simple but effective method. Furthermore, compared with the learnable classifier, the fixed classifier avoid overfitting to previous tasks, which prevents catastrophic forgetting implicitly.

Using the fixed random classifier will leads a problem: the random seed makes the performance of fixed classifiers unstable. To solve this problem, we propose a parameter rearrangement algorithm which implicitly reduces the latent representations drift in old tasks. Suppose $\displaystyle \hat{\vw}^{\text{A}}$ and $\displaystyle \hat{\vw}^{\text{B}}$ are equivalent fixed classifiers used in task A and task B respectively, it’s noteworthy that $\displaystyle \hat{\vw}^{\text{A}}$ and $\displaystyle \hat{\vw}^{\text{B}}$ span the subspace of $\displaystyle \rvh_2$; therefore, ensuring $\displaystyle \hat{\vw}^{\text{B}}$ orthogonal to $\displaystyle \hat{\vw}^{\text{A}}$ will reduce the drift. An intuitive idea is to continually sample $\displaystyle \hat{\vw}^{\text{B}}$ until $\displaystyle (\hat{\vw}^{\text{B}})^\top\hat{\vw}^{\text{A}}$ closer to zero, but it is too inefficient. Besides, we hope that the norm of 
$\displaystyle \hat{\vw}^{\text{B}}$ will not increase. Therefore, we propose the method of rearranging the entries of $\displaystyle \hat{\vw}^{\text{B}}$ to make $\displaystyle (\hat{\vw}^{\text{B}})^\top\hat{\vw}^{\text{A}}$ approach zero. The advantage of the rearranging algorithm is that it neither affects the probability distribution of the classifier $\displaystyle \vw^{\text{B}}$ nor changes the norm of $\displaystyle \hat{\vw}^{\text{B}}$.

\begin{figure}[h]
    \centering
    \includegraphics[scale=0.4]{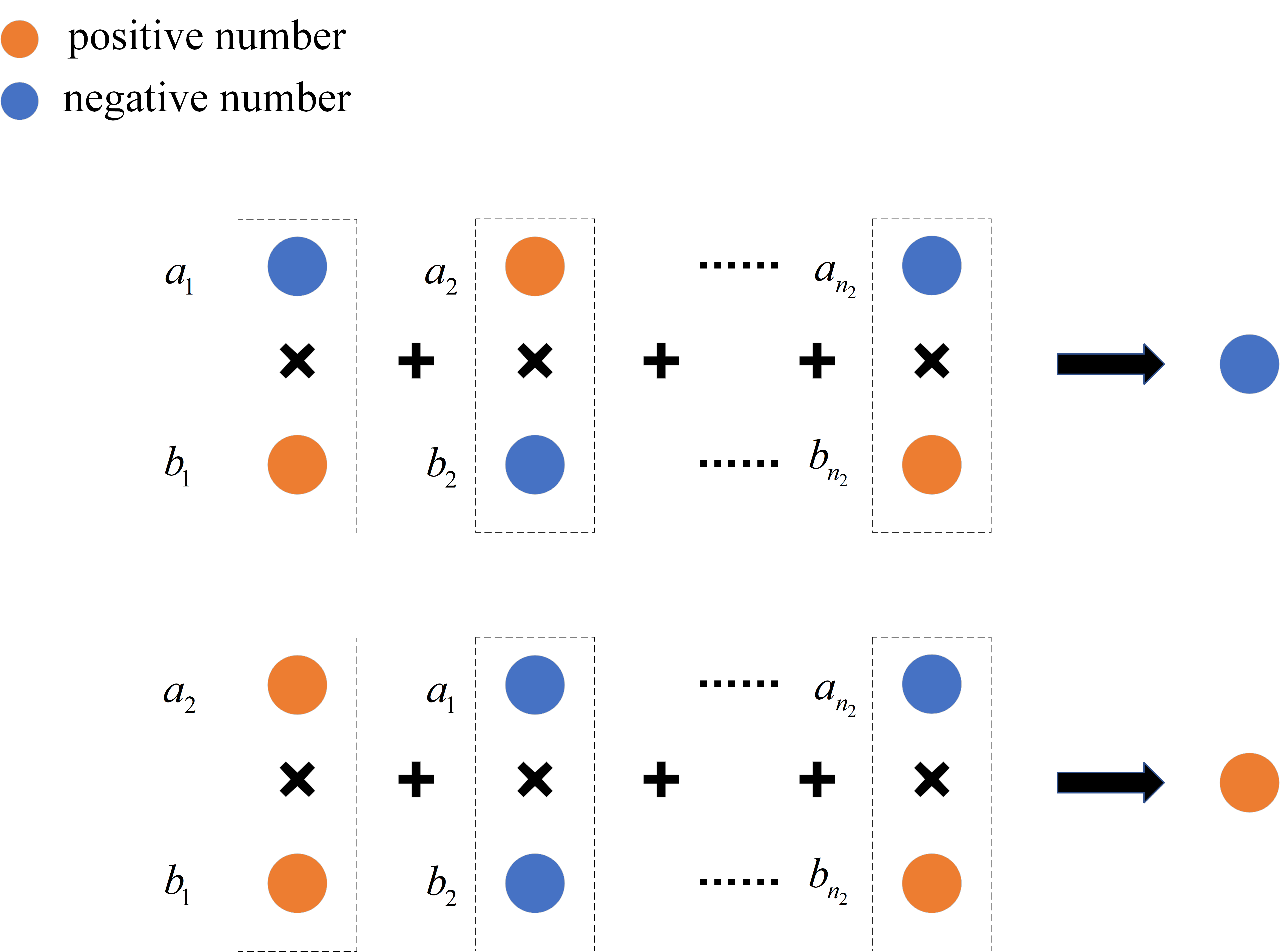}
    \caption{Classifier Entries Rearrangement Algorithm}
    \label{fig2}
\end{figure}

The details of the FRCR algorithm are described below. As shown in Fig.\ref{fig2}, let $\displaystyle \hat{\vw}^{\text{A}}=\left[a_1,...,a_{n_2}\right]^\top, \hat{\vw}^{\text{B}}=\left[b_1,...,b_{n_2}\right]^\top$, then:
\begin{equation}\label{eq9}
\left(\hat{\vw}^{\text{B}}\right)^\top\hat{\vw}^{\text{A}}=\sum_{i=1}^{n_2}a_ib_i
\end{equation}
If $\displaystyle \left(\hat{\vw}^{\text{B}}\right)^\top\hat{\vw}^{\text{A}} < 0$ and $\displaystyle a_1,a_2<0, b_1,b_2>0$, then $\displaystyle \left| \left(\hat{\vw}^{\text{B}}\right)^\top\hat{\vw}^{\text{A}} \right|$ can be probably minimized by swapping the positions of $\displaystyle a_1$ and $\displaystyle a_2$, provided that $\displaystyle n_2$ is large enough. By exchanging all the entries that satisfy the exchange condition one by one, and saving the results of each exchange, we can eventually obtain an answer that minimizes $\displaystyle \left| \left(\hat{\vw}^{\text{B}}\right)^\top\hat{\vw}^{\text{A}} \right|$.
The details of the FRCR algorithm are given in Algorithm \ref{FRCR} and Algorithm \ref{FRCR-CL}.

\subsection{Theoretical Analysis}\label{section:4.2}
Suppose $\displaystyle \hat{\vw}^{\text{A}}=\vw_1^{\text{A}}-\vw_2^{\text{A}}, \hat{\vw}^{\text{B}}=\vw_1^{\text{B}}-\vw_2^{\text{B}}$. According to the back propagation algorithm:
\begin{equation}\label{eq10}
\nabla_{\rvh_2} L=
\mW^\top\frac{\partial L} {\partial \rvy}=
\left[\vw_1 \, \vw_2 \right]\left[\frac{\partial L} {\partial \ry_1} \, \frac{\partial L} {\partial \ry_2} \right]^\top
\end{equation}
In the binary classification tasks using cross entropy loss, we have $\displaystyle \nabla_{\ry_1} L = -\nabla_{\ry_2} L$, so Eq.(\ref{eq10}) is rewritten as:
\begin{equation}\label{eq11}
\nabla_{\rvh_2} L=
\frac{\partial L} {\partial \ry_1}\left(\vw_1-\vw_2 \right)=
\frac{\partial L} {\partial \ry_1}\hat{\vw}
\end{equation}
From Eq. (11), we can see that $\displaystyle \nabla_{\rvh_2} L$ is the subspace spanned by $\displaystyle \hat{\vw}$. In other words, the parameters of the equivalent one-class classifier directly affect the training trajectory of the neural network. In other words, constructing a different initialization parameter   can change the direction of latent representation drift.

Now, we mainly consider how the classifier affect the drift of $\displaystyle \rvh_2$. Assuming that the update of parameter $\displaystyle \mV$ at the beginning of task B training procedure is $\displaystyle \Delta\mV$, the expression for $\displaystyle \Delta\mV$ under the SGD algorithm is:
\begin{equation}\label{eq12}
\Delta\mV=\nabla_{\mV} L=-\eta\frac{\partial L} {\partial \rvh_2^{\text{B}(t_{\text{A}}+1)}} \left(\rvh_1^{\text{B}(t_{\text{A}}+1)} \right)^\top
\end{equation}
where $\displaystyle \eta$ is the learning rate, $\displaystyle \rvh_1^{\text{B}(t_{\text{A}}+1)}$ and $\displaystyle \rvh_2^{\text{B}(t_{\text{A}}+1)}$ are latent representations of task B at moment $\displaystyle t_{\text{A}}+1$. Under the premise that $\displaystyle \rvh_1^{\text{A}(t_{\text{A}})}$ is kept constant, the latent representations drift of task A can be expressed as:
\begin{equation}\label{eq13}
\rvh_2^{\text{A}(t_{\text{A}}+1)}-\rvh_2^{\text{A}(t_{\text{A}})}=
\Delta\mV\rvh_2^{\text{A}(t_{\text{A}})}
\end{equation}
Substituting Eq.(\ref{eq11}) and Eq.(\ref{eq12}), Eq.(\ref{eq13}) is rewritten as:
\begin{equation}\label{eq14}
\rvh_2^{\text{A}(t_{\text{A}}+1)}-\rvh_2^{\text{A}(t_{\text{A}})}=
\rr \hat{\vw}^\text{B}
\end{equation}
The expression for $\displaystyle \rr \in \R$ follows:
\begin{equation}\label{eq15}
\rr = -\eta \left(\rvh_1^{\text{B}(t_{\text{A}}+1)} \right)^\top \rvh_1^{\text{A}(t_{\text{A}})} \frac{\partial L}{\partial \ry_1^\text{B}}
\end{equation}
where $\displaystyle \left(\rvh_1^{\text{B}(t_{\text{A}}+1)} \right)^\top \rvh_1^{\text{A}(t_{\text{A}})}$ is the Gram matrix which characterizes the correlation between $\displaystyle \rvh_1^{\text{A}(t_{\text{A}})}$ and $\displaystyle \rvh_1^{\text{B}(t_{\text{A}}+1)}$, and $\displaystyle \partial L / \partial \ry_1^\text{B}$ is the error signal of the first component of $\displaystyle \rvy^\text{B}$.
Combining Eq.(\ref{eq5}), Eq.(\ref{eq6}) and Eq.(\ref{eq14}), we obtain:
\begin{equation}\label{eq16}
    \begin{aligned}
        \Delta \hat{y}^\text{A}
        &= \left| \left(\hat{\vw}^\text{A} \right)^\top \left(\rvh_2^{\text{A} (t_\text{B})} - \rvh_2^{\text{A}(t_\text{A})}\right) \right| \\
        &= \left| \rr \right| \left| \left(\hat{\vw}^\text{A} \right)^\top \hat{\vw}^\text{B} \right|
    \end{aligned}
\end{equation}
According to Eq.(\ref{eq16}), the equivalent prediction offset consists of two terms, the first term $\displaystyle \rr$ is the product of the Gram matrix and the error signal, the second term 
$\displaystyle \left(\hat{\vw}^\text{A} \right)^\top \hat{\vw}^\text{B}$ represents the correlation between $\displaystyle \hat{\vw}^\text{A}$ and $\displaystyle \hat{\vw}^\text{B}$ . Therefore, decreasing $\displaystyle \left| \left(\hat{\vw}^\text{A} \right)^\top \hat{\vw}^\text{B} \right|$ can reduce the variation of equivalent prediction and mitigate the catastrophic forgetting.

\section{Experiments}
\subsection{Dataset}
\textbf{5-Split-MNIST}: The MNIST dataset\citep{RN33} contains 70,000 grayscale images of handwritten digits of size 28x28, of which 60,000 are used in the training dataset and 10,000 are used in the test dataset. The 5-Split-MNIST dataset divides the MNIST dataset into five subsets, each subset containing two classes of digits.

\textbf{5-Split-FashionMNIST}: The Fashion-MNIST dataset \citep{RN34} contains 10 categories of clothing grayscale images. The size of each image is 28x28. There are 60,000 images for the training set and 10,000 images for the test dataset. The 5-Split-FashionMNIST divides Fashion-MNIST into five subsets, each subset contains two categories.

\textbf{5-Split-CIFAR10}: The CIFAR-10 dataset\citep{RN35} contains 60,000 32x32 color images with 10 categories. There are 50,000 images for the training set and 10,000 images for the test set. 5-Split-CIFAR10 divides the CIFAR-10 dataset into 5 subsets, each subset contains two categories.

\subsection{Metrics}
\textbf{Final average accuracy.} This metric represents the average accuracy of the model on the first $\displaystyle T$ tasks after the task $\displaystyle T$ finishes training, and it reflects the effect of the current backbone on past tasks after the training of task $\displaystyle T$ is completed:
\begin{equation}\label{eq17}
\text{ACC} = \frac{1}{T}\sum_{i=1}^{T} a_{T,i}
\end{equation}
where $\displaystyle a_{T,i}$ denotes the accuracy of the task $\displaystyle i$ after the task $\displaystyle T$ finishes training.

\textbf{Average maximum forgetting rate.} This metric records the maximum decrease in accuracy for any one task $\displaystyle i$ since it finishes training, and averages the maximum decrease over all tasks:
\begin{equation}\label{eq18}
\text{AMF} = \frac{1}{T-1}\sum_{i=1}^{T-1} \max_{j>i} \left(a_{i,i}-a_{j,i}\right)
\end{equation}
where $\displaystyle a_{j,i}$ denotes the accuracy of the task 
$\displaystyle i$ after the task $\displaystyle j$ finishes training.

\subsection{Implements}
The neural network used in the experiments contains two parts: the backbone and the classifier. The backbone is a two-layer hidden layer MLP, each hidden layer contains 256 neurons on the 5-Split-MNIST and 5-Split-FashionMNIST datasets, and each hidden layer contains 2000 neurons on the 5-Split-CIFAR10 dataset. The classifier uses a fully connected layer. The detail of network structure is shown in Appendix \ref{Appendix:B}.

\subsection{Results}
In this paper, experiments were conducted under five random seeds, and the final experimental results are shown in Table \ref{table1}. Among them, FRC stands for fixed random classifier algorithm and FRCR stands for fixed random classifier rearrangement algorithm. EWC are regularization-based continual learning algorithms, they save the parameters of the previous task, so this type of algorithms is generally more advantageous under the premise that the dataset of the past task is not used and the capacity of the network is unchanged. stable SGD is similar to the algorithms proposed in this paper, which neither saves the parameters of the past task nor uses the dataset of the past task. Similar to the algorithm proposed in this paper, stable SGD does not save the parameters of the past tasks and does not use the dataset of the past tasks. From the experimental results, compared with the baseline, the performance of MLP can be improved a lot by just relying on the fixed stochastic classifier, and the indexes are even comparable to those of EWC. Rearranging the classifier entries further improves the performance of the classifier.

\begin{table*}[t]
\caption{Final average accuracy (ACC) and average maximum forgetting rate (AMF) after the final task.}
\label{table1}
\centering
\resizebox{0.96\columnwidth}{!}
{
    \begin{tabular}{l c c c c c c}
    &&&&&&\\[0.0em]
    \toprule
    \multirow{2}{*}{Method}                  &
    \multicolumn{2}{c}{5-Split-MNIST}        &
    \multicolumn{2}{c}{5-Split-FashionMNIST} &
    \multicolumn{2}{c}{5-Split-CIFAR10}\\
    {} & ACC & AMF & ACC & AMF & ACC & AMF\\
    \midrule
    baseline    & 95.85\scalebox{0.8}{$\pm$2.69}
                &  4.83\scalebox{0.8}{$\pm$3.33}
                & 92.60\scalebox{0.8}{$\pm$3.83}
                &  9.75\scalebox{0.8}{$\pm$5.06}
                & 73.50\scalebox{0.8}{$\pm$1.35}
                & 14.33\scalebox{0.8}{$\pm$1.42}
                \\
    stable SGD  & 95.98\scalebox{0.8}{$\pm$0.60}
                &  6.25\scalebox{0.8}{$\pm$2.30}
                & 84.07\scalebox{0.8}{$\pm$2.11}
                & 20.09\scalebox{0.8}{$\pm$2.36}
                & 68.20\scalebox{0.8}{$\pm$0.45}
                & 23.56\scalebox{0.8}{$\pm$1.43}
                \\ 
    EWC         & 96.17\scalebox{0.8}{$\pm$2.73}
                &  4.49\scalebox{0.8}{$\pm$3.34}
                & 93.30\scalebox{0.8}{$\pm$2.73}
                &  8.66\scalebox{0.8}{$\pm$3.34}
                & 74.70\scalebox{0.8}{$\pm$1.52}
                & 12.77\scalebox{0.8}{$\pm$1.56}
                \\
    FRC         & 98.45\scalebox{0.8}{$\pm$1.08}
                &  1.41\scalebox{0.8}{$\pm$1.31}
                & 96.06\scalebox{0.8}{$\pm$1.63}
                &  4.65\scalebox{0.8}{$\pm$3.00}
                & 81.85\scalebox{0.8}{$\pm$1.08}
                &  4.69\scalebox{0.8}{$\pm$1.41}
                \\
    FRCR        &\textbf{98.84}\scalebox{0.8}{$\pm$0.19}
                &\textbf{ 0.95}\scalebox{0.8}{$\pm$0.26}
                &\textbf{97.99}\scalebox{0.8}{$\pm$1.15}
                &\textbf{ 2.10}\scalebox{0.8}{$\pm$1.34}
                &\textbf{82.89}\scalebox{0.8}{$\pm$0.26}
                &\textbf{ 3.58}\scalebox{0.8}{$\pm$0.48}
                \Tstrut \\
    \bottomrule
    \end{tabular}
}
\end{table*}

\subsection{Analysis}
\subsubsection{Prediction clustering under learnable and fixed classifiers}
\begin{figure}[h]
    \centering
    \includegraphics[scale=0.16]{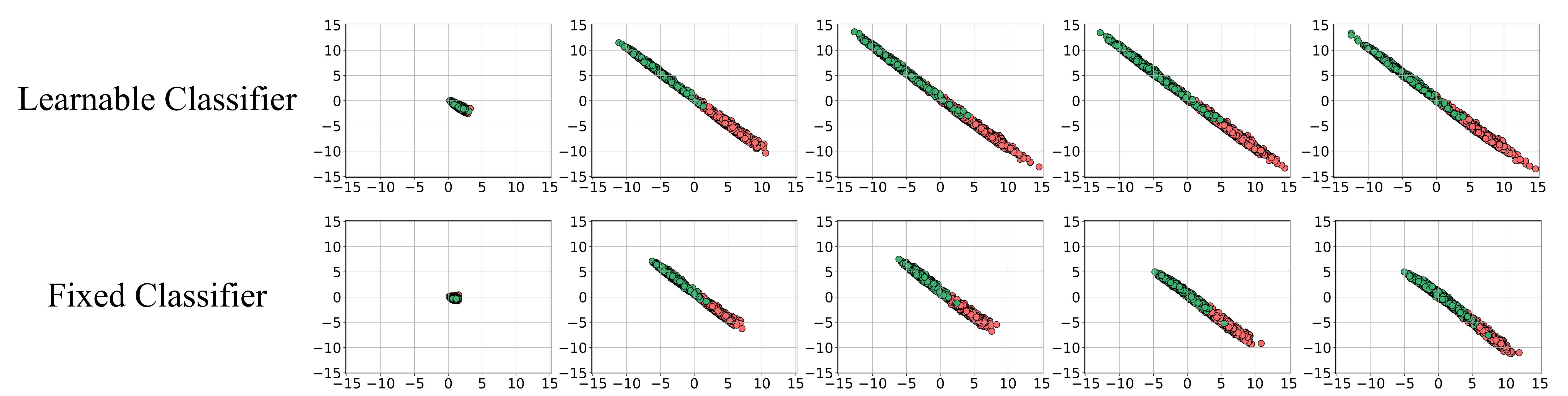}
    \caption{Clustering results of learnable classifiers versus fixed classifiers}
    \label{fig3}
\end{figure}

In order to visualize the effects of the learnable and fixed classifiers on the network stability, the clustering results of the predictions of task 2 at five moments are recorded and plotted in Fig.\ref{fig3}, where the red dots represent category 2 and green dots represent category 3. Comparing the two sets of results, we can find that the clustering results of the learnable classifier are much more elongated than those of the fixed classifier, i.e., the variance of the predictions of the learnable classifier is significantly larger and thus more sensitive to the backbone perturbation, while the clustering variance of the fixed classifier does not change much because its parameter paradigm is always the same. That is to say, the variance of the predictions of the learnable classifier is significantly larger and thus more sensitive to the backbone perturbation, while the clustering variance of the fixed classifier does not change much due to the fact that the number of parameters of the fixed classifier is always the same.

\subsubsection{Relevance of latent representations}
\begin{figure}[h]
    \centering
    \includegraphics[scale=0.193]{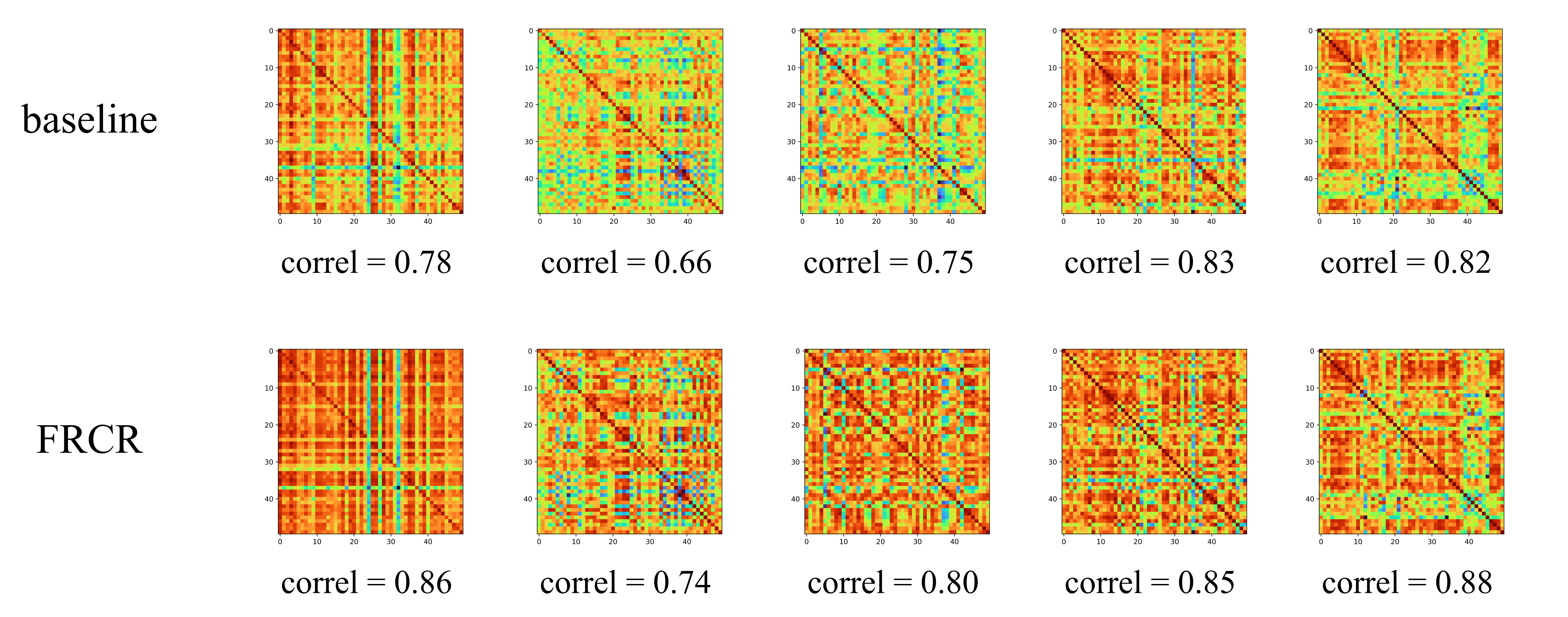}
    \caption{Correlation of latent representations $\displaystyle \rvh_1$ of category 2 images after all tasks finish.}
    \label{fig4}
\end{figure}

As mentioned in Section \ref{section:4.2}, the subspace of $\displaystyle \nabla_{\rvh_2} L$ is formed by $\displaystyle \hat{w}$. Therefore, the subspace of $\displaystyle \nabla_{\rvh_2} L$ remains unchanged during training after freezing the classifier. Since the reordering algorithm makes $\displaystyle \hat{\vw}^{\text{A}}$ and $\displaystyle \hat{\vw}^{\text{B}}$ orthogonal, the correlation between $\displaystyle \rvh_2 ^{\text{A}}$ and $\displaystyle \rvh_2 ^{\text{B}}$ should be very low in theory. On this basis, we find a more interesting phenomenon. As shown in Fig.\ref{fig4}, the correlation of the latent representation $\displaystyle \rvh_1 ^{c_2}$ of category 2 increases after orthogonalizing the equivalent one-class classifiers, which indicates that the degree of $\displaystyle \rvh_1 ^{c_2}$ bias decreases, i.e., the training of the subsequent tasks has less impact on $\displaystyle \rvh_1 ^{c_2}$.

\section{Conclusion}
In this paper, we first analyze the prediction offsets of three-layer MLPs in a continual learning classification task scenario using a visualization method, and find that the degree of forgetting of the neural network is related to the norm of the equivalent one-class classifiers. Based on this conclusion, this paper proposes the use of a fixed random classifier, which can constrain the norm of the equivalent one-class classifier without affecting the performance of the network. In order to ensure the quality of the fixed random classifier, a classifier parameter rearrangement algorithm is proposed, which reduces the correlation between the equivalent one-class classifiers of neighboring tasks by exchanging the entries of the classifier parameters in order to ensure the orthogonality of the latent representations of the different tasks, and thus avoiding the interference of the training of the succeeding tasks with the previous task. Finally, this paper conducts several experiments on 5-Split-MNIST dataset, 5-Split-FashionMNIST dataset and 5-Split-CIFAR10 dataset to verify the effectiveness of the algorithm, and the subsequent experimental analysis shows that the FRCR algorithm indeed mitigates the catastrophic forgetting phenomenon.

The fixed random classifier parameter rearrangement algorithm is a simple, lightweight and efficient continual learning algorithm, which has the advantage that it does not need extra space to store data samples or synaptic importance weights, nor does it need to spend a lot of time to compute regular terms or screen data samples. However, since the design of the fixed random classifier parameter rearrangement algorithm is motivated by the analysis results of the classifiers on three-layer MLP, the algorithm has a number of limitations: (1) the algorithm is only applicable to continual learning scenarios for binary classification tasks; (2) the effectiveness of the algorithm has only been verified on two-layer perceptron machines; and (3) the algorithm's performance is unsatisfactory on the deep network. Although the fixed random classifier parameter rearrangement algorithm has many limitations at present, the design concept of the algorithm is highly interpretable and provides a new idea for subsequent research.

\bibliography{collas2024_conference}
\bibliographystyle{collas2024_conference}

\appendix
\section{PSEUDO-CODE}
Suppose there are $T$ tasks, each task trains $K$ epochs, $\displaystyle \hat{\vw}^{\text{A}}=\vw_1^\text{A}-\vw_2^\text{A}=\left[a_1,...,a_{n_2}\right]^\top$,$ \hat{\vw}^{\text{B}}=\vw_1^\text{B}-\vw_2^\text{B}=\left[b_1,...,b_{n_2}\right]^\top$, the loss function is cross-entropy loss.
\begin{algorithm}[H]
\caption{Fixed random classifier rearrangement (FRCR)}\label{FRCR}
    \begin{algorithmic}[1]
        \Require Classifier $\mW^B=\left[w^\text{B}_1 \, w^\text{B}_2 \right]^\top$,
        classifier $\mW^\text{A}$\\ $\mW_\text{opt}^\text{B} \gets \mW^\text{B}$\\
        $c_{opt} \gets \left| \left(\hat{\vw}^{\text{B}}\right)^\top \hat{w}^\text{A} \right|, c \gets \left| \left(\hat{\vw}^{\text{B}}\right)^\top \hat{w}^\text{A} \right|$\\
        $s_1=\left\{ i|a_i<0,b_i<0 \right\}$,$s_2=\left\{ i|a_i>0,b_i>0 \right\}$ //Save the indices satisfying condition \\
        $d_1=\left\{ i|a_i<0,b_i>0 \right\}$,$d_2=\left\{ i|a_i>0,b_i<0 \right\}$\\
        $k_1 = \min \left(\left|s_1\right|,\left|s_2\right|\right)$, $k_2 = \min \left(\left|s_1\right|,\left|s_2\right|\right)$
        $m_1 \gets 0$, $m_2 \gets 0$
        \Statex
        \Repeat
            \If {$c<0$}
                \State $i=s_1[m_1], j=s_2[m_1], m_1 \gets m_1+1$
            \Else
                \State $i=d_1[m_2], j=d_2[m_2], m_2 \gets m_2+1$
            \EndIf
            \State Exchange the $i$-th column and $j$-th column entries of $\mW^\text{B}$
            \State $c \gets \left|\left(\hat{\vw}^{\text{B}}\right)^\top\hat{\vw}^\text{A}\right|$
            \If {$c<c_\text{opt}$}
                \State $c_\text{opt} \gets c$, $\mW_\text{opt}^\text{B} \gets \mW^\text{B}$
            \EndIf
        \Until {$m_1>k_1$ \textbf{or} $m_2>k_2$}
        \State \Return $\mW_\text{opt}^\text{B}$
    \end{algorithmic}
\end{algorithm}
 
\begin{algorithm}[H]
\caption{Fixed random classifier rearrangement for continual learning (FRCR-CL)}\label{FRCR-CL}
    \begin{algorithmic}[1]
        \Require Training datasets $\left \{\mathcal{D}_t \right\}_{t=1}^T$
        \State Initialize weight $\theta^1=\left\{\theta_b,\mW^1\right\}$, where $\theta_b=\left\{\mU, \mV\right\}$ is backbone, $\mW^1$ is classifier of task 1
        \State Freeze classifier $\mW^1$
        \Statex
        \For{$t=1,...,T$}
            \For{$k=1,...,K$}
                \State Train network $\theta^t$ using dataset $\mathcal{D}_t$
            \EndFor
            \State Generate and randomly initialize new classifier $\mW^{t+1}$
            \State $\mW^{t+1}_\text{opt} \gets$ \hyperref[FRCR]{FRCR}$(\mW^{t+1},\mW^t)$
            \State Freeze classifier $\mW^{t+1}_\text{opt}$
            \State $\theta^{t+1}=\left\{\theta_b,\mW^{t+1}_\text{opt} \right\}$
        \EndFor
    \end{algorithmic}
\end{algorithm}

\newpage
\section{Neural Network Architecture}\label{Appendix:B}
\begin{table*}[h]
\caption{MLP Architecture (5-Split-MNIST and 5-Split-FashionMNIST)}
\label{table2}
\centering
\resizebox{0.96\columnwidth}{!}
{
    \begin{tabular}{c c c}
    &&\\[0.0em]
    \toprule
    {} & Hidden Layer Size & Activation Function\\
    \midrule
    Input Image (784x1) &     &      \\
           FC Layer     & 256 & ReLU \\
           FC Layer     & 256 & ReLU \\
    Task1: FC Layer     &   2 &      \\
    $\cdots$            &   2 &      \\
    Task5: FC Layer     &   2 &      \\
    \bottomrule
    \end{tabular}
}
\end{table*}

\begin{table*}[h]
\caption{MLP Architecture (5-Split-CIFAR10)}
\label{table3}
\centering
\resizebox{0.96\columnwidth}{!}
{
    \begin{tabular}{c c c}
    &&\\[0.0em]
    \toprule
    {} & Hidden Layer Size & Activation Function\\
    \midrule
    Input Image (3072x1) &      &      \\
           FC Layer      & 2000 & ReLU \\
           FC Layer      & 2000 & ReLU \\
    Task1: FC Layer      &    2 &      \\
    $\cdots$             &    2 &      \\
    Task5: FC Layer      &    2 &      \\
    \bottomrule
    \end{tabular}
}
\end{table*}

\end{document}